\pdfoutput=1

\documentclass[11pt]{article}

\usepackage[]{acl}

\usepackage{times}
\usepackage{latexsym}

\usepackage[T1]{fontenc}

\usepackage[utf8]{inputenc}

\usepackage{microtype}

\usepackage{inconsolata}

\usepackage{graphicx}
\usepackage{natbib}
\usepackage{caption}
\usepackage{algorithm}
\usepackage{listings}
\usepackage{algorithmic}
\usepackage{amsmath}
\usepackage{booktabs}
\usepackage{multirow}
\usepackage[outdir=./]{epstopdf}
\usepackage{enumitem}
\usepackage{caption}
\usepackage{subcaption}
\usepackage{subfloat}
\usepackage{newfloat}
\usepackage{graphicx}
\usepackage{svg} %
\usepackage[normalem]{ulem}
\usepackage{framed}
\usepackage{mdframed}
\usepackage{xcolor}
\usepackage{lipsum}
\usepackage{float}
\usepackage{hyperref}
\usepackage[utf8]{inputenc}
\usepackage{forest}
\usepackage{amssymb}
\usepackage{framed}
\usepackage{xcolor}
\usepackage{listings}
\usepackage[most]{tcolorbox}
\definecolor{shadecolor}{gray}{0.9}

\title{AutoQual: An LLM Agent for Automated Discovery of\\Interpretable Features for Review Quality Assessment}

\author{Xiaochong Lan$^{1,2}$ \ \ \ Jie Feng$^1$\thanks{Corresponding author.} \ \ \ Yinxing Liu$^2$ \ \ \  Xinlei Shi$^2$ \ \ \ Yong Li$^1$\footnotemark[1]\\
$^1$Department of Electronic Engineering, BNRist, Tsinghua University \ \ $^2$Meituan\\
\small \texttt{lanxc22@mails.tsinghua.edu.cn} \ \ \ \texttt{fengjie@tsinghua.edu.cn} \ \ \ \texttt{liyong07@tsinghua.edu.cn}}

\begin{document}
\maketitle

\begin{abstract}
Ranking online reviews by their intrinsic quality is a critical task for e-commerce platforms and information services, impacting user experience and business outcomes. However, \textit{quality} is a domain-dependent and dynamic concept, making its assessment a formidable challenge. Traditional methods relying on hand-crafted features are unscalable across domains and fail to adapt to evolving content patterns, while modern deep learning approaches often produce black-box models that lack interpretability and may prioritize semantics over quality. To address these challenges, we propose AutoQual, an LLM-based agent framework that automates the discovery of interpretable features. While demonstrated on review quality assessment, AutoQual is designed as a general framework for transforming tacit knowledge embedded in data into explicit, computable features. It mimics a human research process, iteratively generating feature hypotheses through reflection, operationalizing them via autonomous tool implementation, and accumulating experience in a persistent memory. We deploy our method on a large-scale online platform with a billion-level user base. Large-scale A/B testing confirms its effectiveness, increasing average reviews viewed per user by 0.79\% and the conversion rate of review readers by 0.27\%. 
\end{abstract}

\section{Introduction} 
Online reviews profoundly influence consumer decisions~\citep{chevalier2006effect, floyd2014online} on platforms like Yelp, Amazon, and Meituan. Consequently, ranking these reviews by their intrinsic \textit{quality} or \textit{helpfulness} is a task of paramount importance~\cite{liu2008modeling}. An effective review ranking system enhances user trust, facilitates informed choices, and ultimately drives business conversions. 

However, quantifying review quality presents two significant challenges. First, quality is highly domain-dependent; criteria for a helpful restaurant review differ substantially from those for a useful product review. Manually engineering feature sets for numerous domains is unscalable. Second, quality is dynamic, as user expectations and content patterns evolve, requiring constant adaptation. 

Existing paradigms struggle to address this complexity. Traditional methods that rely on hand-crafted features are rigid and fail to adapt to new domains or evolving quality standards without manual re-engineering~\citep{kim2006automatically,diaz2018modeling}. While modern deep learning models avoid this manual effort~\cite{fan2019product,chen2019multi}, they often function as uninterpretable black boxes~\citep{rudin2019stop}, hindering diagnostics and offering no actionable insights. Moreover, pre-trained language models (PLMs)~\cite{vaswani2017attention,devlin2019bert} are typically optimized for semantics, not textual quality, making them prone to shortcut learning~\cite{zhou2024llm} and thus suboptimal for this task. 
\begin{figure*}[ht] 
\centering 
\includegraphics[width=0.95\textwidth]{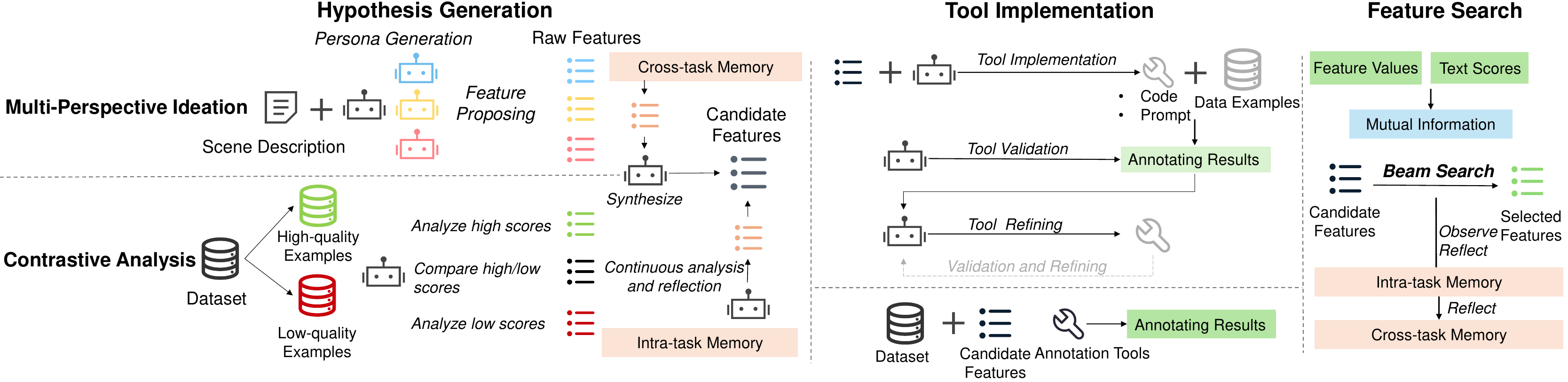} 
\vspace{-0.3cm} 
\caption{AutoQual is an autonomous LLM agent framework for interpretable feature discovery. It operates through hypothesis generation, tool implementation, and reflective search, guided by a dual-level memory.} 
\label{fig:main} 
\vspace{-0.5cm} 
\end{figure*} 

This reveals a critical research gap: the need for a framework that can autonomously discover interpretable and effective features for review quality assessment. To bridge this gap, we propose \textbf{AutoQual}, an autonomous LLM agent framework designed for the automated discovery of interpretable features. AutoQual transforms the tacit knowledge embedded in labeled data into explicit, computable, and interpretable features. It mimics a human research workflow through an iterative cycle: it first \textit{hypothesizes} potential features using multi-perspective ideation and contrastive data analysis; it then \textit{operationalizes} these features by autonomously creating measurement tools (e.g., prompts or scripts); finally, it employs a reflective search guided by a dual-level \textit{memory} system to identify an optimal and compact feature set. AutoQual transforms feature engineering from a manual, ad-hoc process into a scalable, automated operation. 

We demonstrate the real-world effectiveness of AutoQual through its deployment in the review ranking system of a large-scale online platform with a billion-level user base. Large-scale A/B testing confirmed its effectiveness, increasing average reviews viewed per user by 0.79\% and the conversion rate of review readers by 0.27\%. 

Our contributions are: 
\begin{itemize}[nosep,leftmargin=*] 
\item To the best of our knowledge, we are the first to tackle the challenge of discovering interpretable quality features for online reviews by proposing the AutoQual framework. 
\item We propose AutoQual, an LLM-based agent that integrates reflection, tool implementation, and memory to effectively navigate the feature space and discover high-quality, interpretable features. 
\item We demonstrate the significant real-world impact of our method through a large-scale industrial deployment, providing strong evidence of its effectiveness and bridging the gap between academic research and industrial practice. 
\item AutoQual can be viewed as a general framework for transforming the tacit knowledge embedded in expert annotations of unstructured data into explicit, computable, and interpretable features, applicable to a wide range of downstream tasks. We demonstrate its effectiveness across a series of other tasks beyond review quality assessment.
\end{itemize} 

\section{Problem Formulation} 
We focus on the task of review quality assessment, which aims to predict a numerical score for a given review text, where a higher score denotes higher \textit{quality}. 
These scores are often derived from user engagement signals, such as explicit helpfulness votes or implicit signals like click-through rates. Our work focuses on moving beyond black-box predictors by automatically discovering a set of interpretable features whose values are highly predictive of these quality scores.
This approach is generalizable to other domains where interpretable features are desired. 

\paragraph{Problem Definition:} Formally, given a dataset of texts $\mathcal{D} = \{(x_i, y_i)\}_{i=1}^N$, where $x_i$ is a text (e.g., a review) and $y_i \in \mathbb{R}$ its associated target score (e.g., quality score), our objective is to find a set of $k$ interpretable feature functions $\mathcal{S}^* = \{f_1, f_2, \ldots, f_k\}$. Each function $f_j$ maps a text $x_i$ to a feature value, $f_j(x_i)$. We seek the feature set whose output values are maximally informative about the target scores. This is achieved by maximizing the mutual information: 

\begin{equation} 
\mathcal{S}^* = \arg\max_{|\mathcal{S}|=k} I(Y; \mathbf{F}_\mathcal{S}) 
\end{equation} 

where $\mathbf{F}_\mathcal{S}$ denotes the collection of values generated by applying the feature functions in $\mathcal{S}$ to the dataset.
By design, each feature function $f_j$ is \textit{interpretable}, with its definition expressible in natural language (e.g., \textit{contains actionable advice}). 

\section{Method} 
\label{sec:method} 
To solve the problem of automatic interpretable feature discovery, we propose \textbf{AutoQual}, which is illustrated in Figure~\ref{fig:main}. The specific design of each component of AutoQual is explained below. Due to space constraints, the specific prompts used are detailed in Section~\ref{sec:prompts}. 
\subsection{Initial Hypothesis Generation} 
\label{ssec:hypothesis_generation} 
To establish a comprehensive initial candidate pool of features, $\mathcal{S}_{\text{cand}}$, the agent employs two complementary strategies. 

First, it performs \textbf{multi-perspective ideation}. The agent prompts the LLM to instantiate distinct expert personas (e.g., a critical user, a product manager) relevant to the task scenario.
Each persona then proposes a set of features based on its unique evaluation criteria, ensuring diversity in the initial hypotheses. 

Second, the agent conducts \textbf{contrastive analysis} to ground the hypotheses in data.
We first sample high- and low-quality reviews from the dataset $\mathcal{D}$ to construct three distinct sets: (1) high-quality only, (2) low-quality only, and (3) a mix of both. We then employ three corresponding prompts to have the LLM identify the common strengths of high-quality reviews, the common flaws of low-quality reviews, and the key differentiators between them. This process yields three sets of feature hypotheses.

Finally, the agent processes the raw outputs from both strategies, de-duplicating and formalizing them into a candidate feature pool, $\mathcal{S}_{\text{cand}}$. 

\subsection{Autonomous Tool Implementation} 
\label{ssec:tool_implementation} 

For each feature hypothesis $f \in \mathcal{S}_{\text{cand}}$, AutoQual must develop a reliable mechanism to quantify it. To this end, the agent autonomously generates an annotation \textbf{tool}. A tool can be either a programmatic function (e.g., a Python script for syntactic analysis) or a precisely engineered LLM prompt for annotation. 

For each feature hypothesis, an LLM first determines the appropriate tool type (``CODE'' or ``PROMPT''). Subsequently, based on the hypothesis and the selected tool type, the LLM generates the corresponding implementation by creating either the code to compute the feature value or the prompt to annotate it.

To ensure the reliability of these tools, the agent can create each one through an iterative \texttt{propose-validate-refine} cycle. In this process, a newly generated tool is immediately validated on a small sample of reviews. We use a dedicated ``validate tool'' prompt, providing the sample text and the tool's output to the LLM, which then decides if refinement is needed. If the performance is deemed inadequate, the agent refines the tool's logic or prompt. The tool is finalized for subsequent use once the LLM deems its performance satisfactory or a maximum number of refinement cycles is reached. Once finalized, the implemented tool is used to annotate the entire dataset, generating feature values for each review.

\subsection{Reflective Feature Search} 
\label{ssec:reflective_search} 

With the annotated candidate features, AutoQual performs a reflective search to find the optimal set $\mathcal{S}^*$. The search is structured as a \textbf{beam search} algorithm with a beam width of $m$ to balance exploration and computational cost. 

The search is initialized by selecting the top $m$ features with the highest mutual information with the target scores, $I(Y; \mathbf{f}_j)$, to form the initial beams. Here, $\mathbf{f}_j$ represents the vector of values generated by applying feature function $f_j$ to the dataset. Next, through beam expansion and reflection, the agent selects the optimal feature set. 

\paragraph{Beam Expansion.} 
For each of the $m$ beams beam, representing a current feature set $\mathcal{S}_{\text{current}}$, the agent selects the next feature $f_{\text{new}}$ from the updated candidate pool that maximizes the conditional mutual information: 
\begin{equation} 
f^* = \arg\max_{f_{\text{new}} \in \mathcal{S}_{\text{cand}} \setminus \mathcal{S}_{\text{current}}} I(Y; \mathbf{f}_{\text{new}} | \mathbf{F}_{\mathcal{S}_{\text{current}}}) 
\end{equation} 
where $\mathbf{F}_{\mathcal{S}_{\text{current}}}$ denotes the collection of feature values for the set $\mathcal{S}_{\text{current}}$. This ensures that newly added features provide maximal novel information not already captured by the existing set. The process of reflection and expansion repeats until each beam contains $k$ features. The beam with the highest overall joint mutual information $I(Y; \mathbf{F}_{\mathcal{S}})$ is chosen as the final result $\mathcal{S}^*$. 

\paragraph{Intra-Task Reflection and Re-hypothesization.}
After the initial selection of $k$ features, the agent performs intra-task reflection and re-hypothesization, attempting to propose more effective potential features. It takes the current context as input, observing the currently selected feature set, the mutual information between the features and the target metric, and the problem context.
The LLM reflects on the performance of existing features to distill general principles of feature effectiveness for the given domain.
Based on these insights, it then generates new hypotheses designed to be more predictive or to capture novel information.
For example, if features related to \textit{concreteness} are found to be effective, the LLM might hypothesize new, more nuanced features like \textit{presence of statistical evidence} or \textit{use of illustrative anecdotes}. Additionally, the agent observes and reflects on the gaps in the current feature coverage dimensions, proposing features that contribute to comprehensive coverage. These new hypotheses are dynamically added to the candidate pool. The beam search is then re-run on this augmented pool to select the optimal set of $k$ features. The agent repeats this cycle of selection, reflection, and pool augmentation for a predetermined number of iterations.

\subsection{Dual-Level Memory Architecture} 
\label{ssec:memory} 
To enable AutoQual to accumulate experience within a single task and transfer knowledge across multiple tasks, we designed a dual-level memory architecture. 

\paragraph{Intra-Task Memory (Working Memory).} 
This memory operates during a single discovery task. It maintains the state of the reflective search, including tested features and their MI scores. Based on this memory, the agent generates intermediate insights through reflection.
This allows the agent to assess the quality of its past decisions and dynamically adapt its strategy within a single problem-solving session. 

\paragraph{Cross-Task Memory (Long-Term Memory).} 
Upon completing a task, the agent synthesizes the problem description, the final feature set $\mathcal{S}^*$, and their mutual information into a consolidated summary. This summary is stored in a persistent, long-term knowledge base. When AutoQual is presented with a new task, it queries this knowledge base for relevant prior experiences. These retrieved experiences serve as an additional, informed source for the initial hypothesis generation (Section~\ref{ssec:hypothesis_generation}), allowing the agent to bootstrap its search and improve its performance over time.

\section{Experimental Setup}
\subsection{Implementation Details}
\label{sec:implementation}
AutoQual is open-sourced\footnote{https://github.com/tsinghua-fib-lab/AutoQual}. For the agent's core components, including hypothesis generation, tool implementation, memory, and reflection, we employ DeepSeek-V3.2-Exp (Thinking Mode) as the backbone LLM.
Its inherent reasoning capabilities allow us to bypass the manual workflow optimization for each individual submodule of our method.
For the more scalable task of feature annotation, we use the cost-effective qwen-plus-latest model~\cite{yang2025qwen3}.
To ensure reproducibility, we set the generation temperature to 0. We estimate mutual information using a KNN-based estimator, leveraging the implementation available in the scikit-learn library~\cite{pedregosa2011scikit}.
In our reflective search, we use a beam width of $m=5$ and select a final set of $k=10$ features. All reported results for our main method, baselines, and ablations are averaged over five independent runs.
To maintain a clean experimental setup, our main experiments do not utilize the cross-task memory; its benefits are analyzed separately in the ablation study.

\subsection{Datasets}
Our primary experiments leverage two review datasets: the public Amazon review dataset~\citep{hou2024bridging} and a private dataset from Meituan.
For Amazon, we sample 2,000 representative reviews from each of four categories (Cellphones and Accessories, Clothing, Shoes and Jewelry, Grocery and Gourmet Food, and Office Products), using \textit{helpful votes} as the quality score.
For Meituan, we sample 20,000 reviews from the \textit{in-store dining} domain, using review \textit{click-through rate (CTR)} as the quality score.
We also utilize additional datasets for versatility testing (Section~\ref{ssec:versatility}).

\subsection{Evaluation Metrics}
Following prior work~\cite{zhou2024llm}, we evaluate the agreement between the predicted and the ground-truth scores for regression tasks (like quality assessment) using two standard metrics:
\paragraph{Spearman’s Rho ($r_s$).} It measures the strength and direction of the monotonic relationship between two ranked variables. Given $N$ pairs of ranks $(R(y_i), R(\hat{y}_i))$ and the difference $d_i = R(y_i) - R(\hat{y}_i)$, it is defined as $r_s = 1 - \frac{6 \sum_{i=1}^{N} d_i^2}{N(N^2 - 1)}$.
\paragraph{Mean Absolute Error (MAE).} It measures the average magnitude of the errors between predicted and actual scores. It is defined as $\text{MAE} = \frac{1}{N} \sum_{i=1}^{N} |y_i - \hat{y}_i|$.

Since the scales of helpfulness votes and CTR differ significantly, we normalize the ground-truth scores to the $[0, 1]$ range before calculating MAE.
For classification tasks (like toxicity detection), we use standard metrics: F1-Score and Area Under the ROC Curve (AUROC).

\subsection{Comparison Methods}

Following prior work~\citep{zhou2024llm}, we compare AutoQual against several well-established baselines. The first group consists of general text modeling approaches: \textbf{Bag-of-Words (BoW)}, which uses TF-IDF features to train a linear regression model; \textbf{Fixed PLM}, which feeds frozen embeddings from a pre-trained language model to a linear regressor; \textbf{Finetuned PLM}, which fine-tunes the PLM end-to-end with a regression head (or classification head for classification tasks); and \textbf{LLM-based methods} in both zero-shot and 20-shot settings for direct scoring. To showcase the full potential of PLM-based approaches, we use a modern embedding model, \texttt{bge-small} (384 dimensions)~\cite{bge_embedding}, which provides stronger representational capacity than older models like BERT, while its moderate dimensionality mitigates the risk of overfitting. For the LLM baselines, we use \texttt{qwen-plus-latest}~\cite{yang2025qwen3}. We also introduce a second group of baselines specifically designed for review helpfulness prediction: \textbf{TNN}~\cite{olmedilla2022prediction}, a 1D-CNN-based model; \textbf{SEHP}~\cite{malik2024sehp}, a stacking-based ensemble model; and \textbf{BHeIP-CoRT}~\cite{li2025bert}, a BERT-based model that utilizes rating-text consistency. While these models are relatively recent, they rely on older backbone embeddings combined with task-specific designs.

It is important to note that our primary goal is to demonstrate the effectiveness of the features discovered by AutoQual. Therefore, for our method and several baselines in the review quality tasks, we deliberately use a simple linear regression predictor. Employing a more sophisticated predictor would likely boost the performance of all methods commensurately, but our focus here is on the inherent predictive power of the features themselves.
\section{Experimental Results}
In this section, we conduct extensive experiments to answer the following four research questions:
\begin{itemize}[nosep,leftmargin=*,itemsep=0pt]
\item\textbf{RQ1:} Can our method effectively discover features that are predictive for the review quality assessment task?
\item\textbf{RQ2:} Does each component of our method contribute to its performance?
\item\textbf{RQ3:} Can our method discover domain-specific, interpretable features?
\item\textbf{RQ4:} Is the AutoQual framework generalizable to other text assessment scenarios?
\end{itemize}

\begin{table*}[t]
\centering
\caption{Performance comparison of AutoQual against baseline methods across different domains. We report Spearman's correlation coefficient ($r_s$, higher is better) and Mean Absolute Error (MAE, lower is better). \textbf{Bold} and \underline{underlined} refer to the best and 2nd best performance.}
\vspace{-0.2cm}
\label{tab:main_results}
\begingroup
\resizebox{0.95\linewidth}{!}{
\begin{tabular}{c|cc|cc|cc|cc|cc}
\hline
\multirow{2}{*}{\textbf{Method}} &
\multicolumn{2}{c|}{\textbf{Meituan}} &
\multicolumn{2}{c|}{\textbf{Cell Phones}} &
\multicolumn{2}{c|}{\textbf{Office Products}} &
\multicolumn{2}{c|}{\textbf{Clothing}} &
\multicolumn{2}{c}{\textbf{Grocery}} \\
& $r_s$ $\uparrow$ & MAE $\downarrow$ & $r_s$ $\uparrow$ & MAE $\downarrow$ & $r_s$ $\uparrow$ & MAE $\downarrow$ & $r_s$ $\uparrow$ & MAE $\downarrow$ & $r_s$ $\uparrow$ & MAE $\downarrow$ \\
\hline
Bag-of-words & 0.3312 & 0.1923 & 0.5269 & 0.0642 & 0.5224 & 0.0445 & 0.5945 & 0.0718 & 0.4124 & 0.0384 \\
Fixed PLM & 0.5023 & 0.1521 & 0.6064 & 0.0566 & 0.6350 & 0.0391 & 0.7355 & 0.0597 & 0.5375 & 0.0349 \\
Fine-tuned PLM & 0.5656 & 0.1229 & 0.6977 & 0.0443 & \underline{0.6934} & 0.0333 & \underline{0.7816} & \underline{0.0467} & 0.6770 & 0.0302 \\
Zero-shot LLM & 0.1237 & 0.2245 & 0.0614 & 0.5274 & -0.0415 & 0.5580 & 0.1315 & 0.5802 & 0.0532 & 0.5588 \\
20-shot LLM & 0.1413 & 0.2124 & 0.0979 & 0.4246 & 0.0459 & 0.4407 & 0.0061 & 0.4955 & 0.0199 & 0.4041 \\
TNN & 0.4141 & 0.1625 & 0.5831 & 0.0612 & 0.5768 & 0.0452 & 0.5121 & 0.0901 & 0.5926 & 0.0349 \\
SEHP & 0.4477 & 0.1521 & 0.5993 & 0.0578 & 0.6072 & 0.0422 & 0.5233 & 0.1014 & 0.6435 & \underline{0.0298} \\
BHelP-CoRT & 0.4919 & 0.1492 & 0.6018 & 0.0586 & 0.5947 & 0.0443 & 0.5435 & 0.0989 & 0.6326 & 0.0312 \\
\hline
AutoQual & \underline{0.5661} & \underline{0.1220} & \underline{0.6992} & \textbf{0.0438} & 0.6646 & \underline{0.0331} & 0.6278 & 0.0662 & \underline{0.7194} & \textbf{0.0261} \\
AutoQual+PLM & \textbf{0.5833} & \textbf{0.1114} & \textbf{0.7105} & \underline{0.0448} & \textbf{0.7189} & \textbf{0.0320} & \textbf{0.8014} & \textbf{0.0451} & \textbf{0.7920} & 0.0475 \\
\hline
\end{tabular}}
\endgroup
\end{table*}
\subsection{Feature Discovery Performance (RQ1)}
We measure the performance of predicting quality scores using the features discovered by AutoQual with a simple linear regressor. We also evaluate a model that combines our features with PLM embeddings (AutoQual+PLM), where the PLM and the linear regressor are trained end-to-end. The results are presented in Table~\ref{tab:main_results}.

We draw the following key conclusions:
\begin{itemize}[nosep,leftmargin=*,itemsep=0pt]
\item\textbf{Discovered quality features have strong predictive power.}
Predictions made solely using the features discovered by AutoQual already demonstrate strong performance. In some cases, the relatively sparse set of features discovered by AutoQual outperforms the high-dimensional semantic features from even a fine-tuned PLM. This suggests that AutoQual identifies high-order quality features that are more predictive than the purely semantic features captured by PLMs.

\item\textbf{Discovered quality features are complementary to PLM embeddings.} The AutoQual+PLM model achieves the best performance on all datasets in terms of $r_s$, significantly outperforming the finetuned PLM alone. This indicates that the high-order quality features discovered by our method are complementary to the fine-grained semantic information provided by PLMs.

\item\textbf{LLM baselines perform poorly.} Both zero-shot and few-shot LLM baselines show poor predictive performance. This demonstrates that the effectiveness of our method stems from our structured agent design, rather than the innate capabilities of the underlying LLM.
\end{itemize}
\begin{table*}[t]
\centering
\caption{Ablation study of AutoQual across different domains.}
\vspace{-0.2cm}
\label{tab:ablation}
\begingroup
\resizebox{\linewidth}{!}{
\begin{tabular}{c|cc|cc|cc|cc|cc}
\hline
\multirow{2}{*}{\textbf{Method}} &
\multicolumn{2}{c|}{\textbf{Meituan}} &
\multicolumn{2}{c|}{\textbf{Cell Phones}} &
\multicolumn{2}{c|}{\textbf{Office Products}} &
\multicolumn{2}{c|}{\textbf{Clothing}} &
\multicolumn{2}{c}{\textbf{Grocery}} \\
& $r_s$ $\uparrow$ & MAE $\downarrow$ & $r_s$ $\uparrow$ & MAE $\downarrow$ & $r_s$ $\uparrow$ & MAE $\downarrow$ & $r_s$ $\uparrow$ & MAE $\downarrow$ & $r_s$ $\uparrow$ & MAE $\downarrow$ \\
\hline
AutoQual & 0.5661 & 0.1220 & 0.6992 & 0.0438 & 0.6646 & 0.0331 & 0.6278 & 0.0662 & 0.7194 & 0.0261 \\
w/o Multi-perspective Ideation & 0.5425 & 0.1246 & 0.6516 & 0.0466 & 0.6109 & 0.0347 & 0.5928 & 0.0689 & 0.7117 & 0.0264 \\
w/o Contrastive Analysis & 0.5401 & 0.1250 & 0.6378 & 0.0470 & 0.6120 & 0.0349 & 0.5266 & 0.0731 & 0.6920 & 0.0263 \\
w/o Intra-task Memory & 0.5502 & 0.1303 & 0.6843 & 0.0440 & 0.6320 & 0.0336 & 0.6133 & 0.0660 & 0.7121 & 0.0266 \\
w/o Intra-task Mem. (+Cross-task) & 0.5602 & 0.1225 & 0.7055 & 0.0432 & 0.6373 & 0.0334 & 0.6164 & 0.0663 & 0.7211 & 0.0262 \\
\hline
\end{tabular}}
\endgroup
\vspace{-0.2cm}
\end{table*}
\subsection{Ablation Study (RQ2)}
In this section, we examine the contribution of each component of our method. The results are detailed in Table~\ref{tab:ablation}. Our core observations are as follows:
\begin{itemize}[nosep,leftmargin=*]
\item \textbf{Hypothesis Generation.} Removing multi-perspective ideation leads to a significant performance drop, with an average $r_s$ decrease of 0.0335. Removing contrastive analysis results in an even larger degradation, with an average $r_s$ drop of 0.0537. Even subsequent reflection cannot fully compensate for the lack of diverse perspectives in the initial feature proposal stage. This validates the effectiveness of our two-pronged hypothesis generation strategy.

\item \textbf{Dual-Level Memory.} Removing the intra-task memory (which disables reflection) impairs the effectiveness of the discovered features, with an average $r_s$ decrease of 0.0170. This highlights the importance of learning from observation and experience. Interestingly, when we remove the intra-task memory but incorporate cross-task memory accumulated from the other four domains, the model's performance is comparable to the full AutoQual model. This suggests that experience is generalizable across tasks. Furthermore, this cross-task setting without intra-task memory significantly reduces computational costs, decreasing the agent's LLM token consumption by 44.95\% and the annotation LLM token consumption by 29.79\% on average.
\end{itemize}
\subsection{Case Study (RQ3)}
\begin{figure}[t]
\centering
\includegraphics[width=\linewidth]{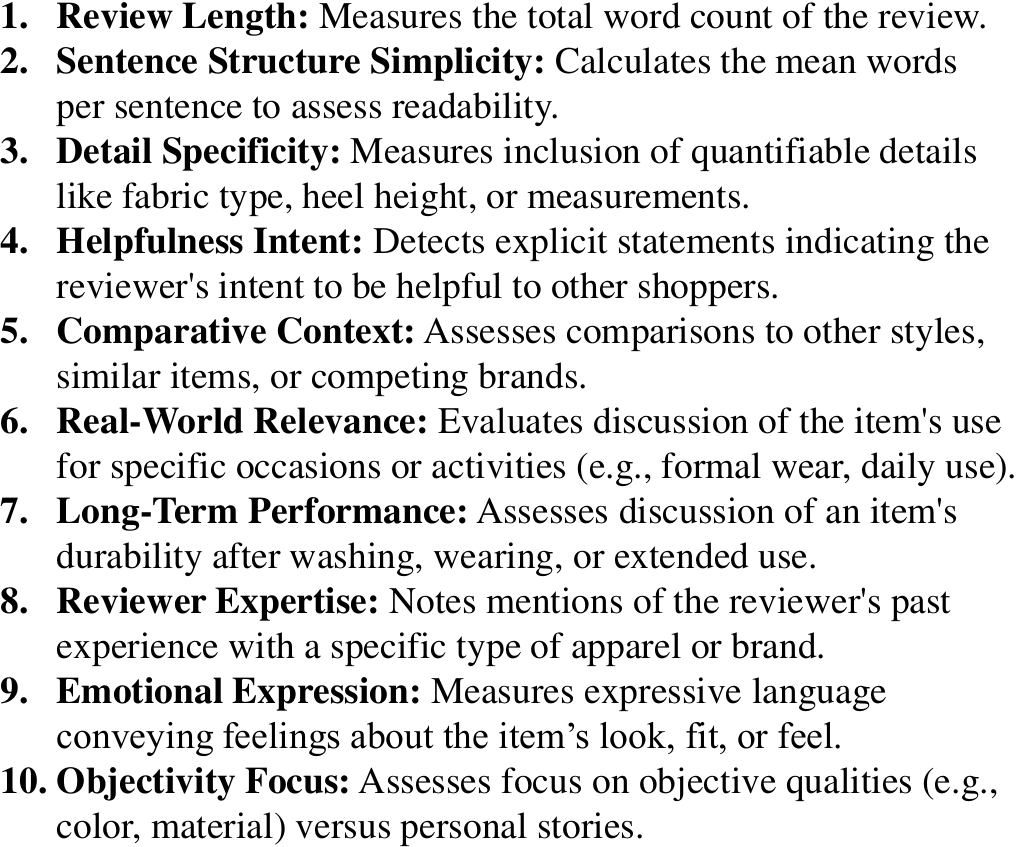}
\vspace{-0.4cm}
\caption{The selected top 10 features in the \textit{Clothing, Shoes, and Jewelry} domain.}
\vspace{-0.5cm}
\label{fig:features}
\end{figure}

We showcase the interpretable features discovered by AutoQual in the \textit{Clothing, Shoes, and Jewelry} domain. The selected top 10 features and their descriptions are shown in Figure~\ref{fig:features}.

\begin{figure}[t]
\centering
\includegraphics[width=1\linewidth]{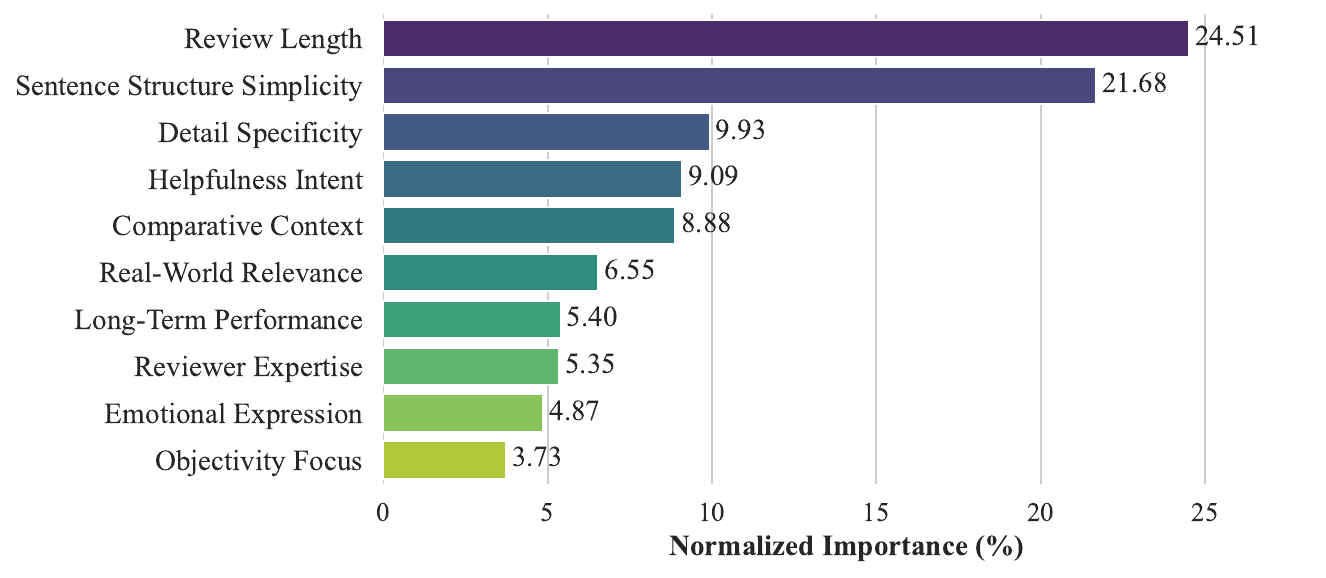}
\vspace{-0.4cm}
\caption{Normalized importance of features discovered by AutoQual in the \textit{Clothing, Shoes, and Jewelry} domain. Importance is measured by mutual information with the quality score.}
\vspace{-0.5cm}
\label{fig:case_study}
\end{figure}

Figure~\ref{fig:case_study} visualizes the normalized importance of these features. As the list demonstrates, many of the discovered features are highly domain-specific (e.g., \textit{Detail Specificity}, \textit{Comparative Context}, \textit{Emotional Expression}). The implicit embeddings from a PLM cannot explicitly capture these effective, high-order features, which explains why its performance lags behind our method.

Furthermore, these interpretable features facilitate model diagnostics. For instance, a feature that is predictive but normatively undesirable can be explicitly identified and removed. They also provide clear guidelines for users on how to write high-quality reviews, which can improve the overall helpfulness of content on the platform. In turn, more helpful reviews help users make informed decisions and can ultimately drive conversions.
\subsection{Generalizability of AutoQual (RQ4)}
\label{ssec:versatility}
We investigate the generalizability of the AutoQual framework by applying it to text assessment tasks beyond review quality. We test our method's capabilities in three additional domains.

First, we assess text persuasiveness on the OUM dataset~\citep{farag2023opening}, following the experimental setup of~\citet{zhou2024llm}. Second, we evaluate automated essay scoring on a subset of the ASAP dataset\footnote{https://www.kaggle.com/competitions/asap-aes}. The results for these tasks, presented in Table~\ref{tab:versatility_regression}, show that the features discovered by AutoQual are highly effective, achieving performance comparable to strong baselines like fine-tuned PLMs and manual feature engineering\footnote{For the persuasiveness task, the manually engineered features are from~\citet{zhou2024llm}. This baseline is omitted for essay scoring due to the lack of a widely-accepted SOTA feature set.}.

\begin{table}[t]
\centering
\caption{Evaluation of AutoQual on regression tasks (Quality/Scoring). The best results are shown in \textbf{bold}.}
\vspace{-0.2cm}
\label{tab:versatility_regression}
\begingroup
\resizebox{1\linewidth}{!}{
\begin{tabular}{c|cc|cc}
\hline
\multirow{2}{*}{\textbf{Method}} &
\multicolumn{2}{c|}{\textbf{Persuasiveness}} &
\multicolumn{2}{c}{\textbf{Essay}} \\
& $r_s$ $\uparrow$ & MAE $\downarrow$ & $r_s$ $\uparrow$ & MAE $\downarrow$ \\
\hline
Finetuned PLM & 0.435 & \textbf{1.395} & \textbf{0.548} & 2.287 \\
Zero-shot LLM & 0.298 & 1.921 & 0.361 & 2.634 \\
20-shot LLM & 0.315 & 2.012 & 0.382 & 2.578 \\
Manual Engineering & \textbf{0.447} & 1.421 & - & - \\
AutoQual & 0.432 & 1.445 & 0.545 & \textbf{2.215} \\
\hline
\end{tabular}}
\endgroup
\end{table}

Third, we apply AutoQual to the Jigsaw Toxic Classification challenge\footnote{https://www.kaggle.com/c/jigsaw-toxic-comment-classification-challenge}. Unlike quality assessment, toxicity detection is often viewed as a more semantics-focused task. We compare a Finetuned PLM against AutoQual+PLM. As shown in Table~\ref{tab:versatility_classification}, the features discovered by AutoQual (e.g., identifying \textit{Dehumanizing Metaphors} or \textit{Inciting Instructions}) provide complementary signals, significantly improving both F1-Score and AUROC.

\begin{table}[t]
\centering
\caption{Evaluation of AutoQual on classification task (Toxicity Detection). The best results are shown in \textbf{bold}.}
\vspace{-0.2cm}
\label{tab:versatility_classification}
\begingroup
\resizebox{0.75\linewidth}{!}{
\begin{tabular}{c|c|c}
\hline
\textbf{Method} & \textbf{F1-Score} $\uparrow$ & \textbf{AUROC} $\uparrow$ \\
\hline
Finetuned PLM & 0.8224 & 0.9078 \\
AutoQual+PLM & \textbf{0.8364} & \textbf{0.9203} \\
\hline
\end{tabular}}
\endgroup
\vspace{-0.5cm}
\end{table}

These results demonstrate that the AutoQual framework is generally applicable for discovering interpretable features across diverse text assessment scenarios. What's more, AutoQual is not just a tool for text quality, but a \textbf{general framework for automated interpretable feature discovery}. This framework can transform the tacit knowledge of domain experts, embedded within labeled data, into explicit, computable, and interpretable features. Specifically, it is applicable to any task that meets the following criteria:
\begin{itemize}[nosep,leftmargin=*]
    \item \textbf{Unstructured Data:} The task involves inputs like text, images, or audio where features are not obvious. By leveraging multimodal foundation models as a backbone, the agent can be extended to handle diverse data types beyond text.
    \item \textbf{Ambiguous Evaluation Criteria:} The target concept is abstract and multi-dimensional (e.g., ``quality,'' ``risk,'' ``helpfulness''), making manual feature engineering difficult and unscalable.
    \item \textbf{Interpretability Requirement:} The decision-making process requires transparency for diagnostics, user trust, or regulatory compliance.
\end{itemize}
\section{Industrial Deployment}
Our method is deployed in a real-world industrial setting on the Meituan platform, which has a billion-level user base. On the platform's merchant detail pages, we aim to rank high-quality reviews higher in the display order. We construct review quality scores using review click-through rates and apply AutoQual to mine features, identifying five key features: informativeness, providing actionable advice, colloquial expression, containing real examples, and credible and engaging language. We further manually add two additional features: not being promotional copy and not being AI-generated. We integrate these features into the existing review ranking model, achieving significant improvements in user review engagement and conversion. In an online A/B experiment conducted from January 18 to February 7, 2025, we observe a 1.42\% increase in average review browsing time, a 0.79\% increase in the average number of reviews viewed per user, and a 0.27\% increase in the conversion rate of users who viewed reviews. This real-world industrial deployment demonstrates the effectiveness of the features discovered by our method.
\section{Conclusion}
In this study, we propose AutoQual, a general LLM-based agent framework for the automated discovery of interpretable features from unstructured data. AutoQual is designed to transform the tacit knowledge of domain experts, often captured implicitly in labeled data, into a set of explicit, computable, and interpretable features. We demonstrate the power of this framework in the challenging domain of review quality assessment, where its iterative cycle of hypothesis generation, autonomous tool implementation, and reflective search successfully discovered highly predictive features. The framework's generalizability is further confirmed by its strong performance across diverse tasks. Finally, its successful deployment on a billion-user platform confirms the significant practical value of automated, interpretable feature discovery for building trustworthy and effective real-world applications.
\section{Limitations}
Our study, while promising, has several limitations that open up avenues for future research:
\begin{itemize}[nosep,leftmargin=*]
\item \textbf{Application to Semantic Tasks.} While we demonstrated AutoQual's effectiveness on toxicity detection, its versatility could be further explored by applying it to other traditional, semantics-focused NLP tasks, such as stance detection or sentiment analysis. This would allow us to further investigate the extent to which the high-order, interpretable features discovered by our method can provide complementary signals to the dense semantic embeddings from pre-trained language models.
\item \textbf{Expanding Domain Coverage.} Our framework can be enhanced by incorporating a broader range of domains beyond the text-based tasks explored. By leveraging multimodal foundation models as a backbone, the agent could handle diverse data types (e.g., images, audio). 
\item \textbf{Refining Industrial Deployment.} Our current industrial deployment is constrained by system architecture limitations, leading us to integrate only a set of high-level, universal features. A key avenue for future work is to incorporate the domain-specific features discovered by AutoQual into our online ranking models. This involves tailoring feature sets for different business scenarios (e.g., restaurants vs. hotels) to further boost ranking performance across the platform.
\end{itemize}
\section{Ethics Statement}
In this work, we analyze user-generated review data. For the public Amazon dataset, we strictly adhered to its terms of use. For the private Meituan dataset, we obtained explicit consent from users for their data to be used in research. Furthermore, all data was rigorously anonymized before use; we removed all user IDs and scrubbed any personally identifiable information (PII) from the review texts. Our research protocol was reviewed and approved by our institution's ethics review board.

Our research utilized large language model services, specifically Deepseek-V3.2-Exp provided by DeepSeek and Qwen-plus provided by Alibaba. We have strictly adhered to the terms of service and usage policies of both model providers.

We acknowledge that our method could be potentially misused. For example, malicious actors could leverage our framework to engineer low-quality or harmful content to achieve higher visibility in ranking systems. We remain vigilant about such risks and advocate for the responsible use of this technology, urging the community to prevent its abuse.
\section{Acknowledgement}
This work was supported in part by National Natural Science Foundation of China under grant U23B2030, in part by the China Postdoctoral Science Foundation under grant 2024M761670 and GZB20240384, in part by the Tsinghua University Shuimu Scholar Program under grant 2023SM235. This work was also supported by Meituan.
\bibliography{custom}
\clearpage
\appendix
\section{Appendix}
\label{sec:appendix}
\subsection{Related Works}
\subsubsection{LLM Agent}
LLM agents augment large language models with carefully designed or self-adaptive workflows, enabling them to tackle more complex tasks. The proactive decomposition of complex tasks by agents is referred to as planning~\citep{yao2023react, shinn2023reflexion}; explicitly stored context during operation is known as memory~\citep{park2023generative, zhong2024memorybank}; and agents can actively invoke external tools to accomplish specific subtasks~\citep{schick2023toolformer, qin2024tool}. Agents have achieved remarkable success across diverse applications, including code generation and software engineering~\citep{jimenez2023swe, yang2024swe}, scientific discovery and research automation~\citep{wang2023scibench, boiko2023autonomous}, interactive web navigation~\citep{deng2023mind2web, zhou2023webarena}, social simulation~\citep{gao2023s3,gao2024large,piao2025agentsociety} and a wide range of fine-grained, specific tasks~\citep{lan2024stance,lan2025benchmarking,feng2024agentmove,xing2025designing,du2024trajagent}. To the best of our knowledge, we are the first to leverage LLM agents for the task of automatic interpretable feature discovery from unstructured text.
\subsubsection{Automated Feature Mining}
For predictive tasks on tabular data, constructing new features from existing data is crucial, such as semantic binning or feature interactions. However, manual feature engineering for every new scenario is time-consuming and labor-intensive, motivating research into automated feature discovery~\citep{fan2010generalized, kanter2015deep, khurana2018feature, li2023learning}. Recent advances in large language models have introduced powerful tools for automated feature discovery, as LLMs possess semantic understanding of both domains and features~\citep{hollmann2023large, nam2024optimized, abhyankar2025llm}. While automated feature generation for \textit{tabular} data has been extensively explored, to the best of our knowledge, no prior work has addressed the automatic discovery and mining of interpretable features from \textit{unstructured text} data. Our work represents an initial exploration in this direction.
\subsubsection{Review Quality Assessment}
Review helpfulness prediction aims to estimate the perceived value of online reviews to potential readers~\citep{liu2008modeling, kim2006automatically}. In our work, we consider additional quality indicators beyond helpfulness votes (e.g., click-through rates), thus framing the task more broadly as \textit{review quality assessment}. Traditional approaches rely on hand-crafted features~\citep{mudambi2010research,ghose2010estimating,diaz2018modeling}, which are time-consuming to develop and difficult to scale across domains. While deep learning methods have shown promise~\citep{fan2019product, chen2019multi}, they often lack interpretability. Current approaches based on pre-trained language models~\citep{vaswani2017attention,liu2019roberta} primarily capture semantic information rather than explicit quality indicators, making them suboptimal for this task. To the best of our knowledge, we are the first to propose automated feature discovery for this task.

\newtcolorbox{promptbox}{
  breakable,
  enhanced,
  colback=gray!15,
  boxrule=0pt,
  left=10pt, right=10pt, top=10pt, bottom=10pt,
  sharp corners,
  before upper = {\setlength{\parindent}{1em}},
}
\subsection{Prompts for AutoQual}
\label{sec:prompts}
This section details the prompts used by the AutoQual agent, corresponding to the different stages of its workflow as described in Section~\ref{sec:method}.

\subsubsection{Initial Hypothesis Generation}
These prompts are used for the initial feature hypothesis generation phase (Section~\ref{ssec:hypothesis_generation}), which includes multi-perspective ideation and contrastive analysis.

\paragraph{Generate Roles Prompt.} Used to create diverse expert personas for multi-perspective ideation. This prompt is as follows:
\begin{promptbox}
We are undertaking a text quality assessment task. The core of this task is to evaluate the quality of text based on the following scenario.

Scenario Description: 

\texttt{\{scene\_description\}}

Your task is to propose \texttt{\{role\_count\}} distinct virtual evaluator roles, each with a different perspective and unique evaluation criteria, based on the text quality requirements of the above scenario. These roles should be representative and cover multiple dimensions of text quality assessment.

Output a list of exactly \texttt{\{role\_count\}} roles. Each role must be on a new line.
Only output \texttt{\{role\_count\}} lines. For each line, provide only the role's name followed by a comma and a concise description of its core evaluation criteria (around 20 words). Do not include any extra explanations, introductory text, or formatting like "Role 1:".

Example Format:

{[Role Name 1]}, {[Core evaluation criteria description]}

{[Role Name 2]}, {[Core evaluation criteria description]}
\end{promptbox}

\paragraph{Generate Features from Role Prompt.} Used by each persona to suggest features from its unique viewpoint. This prompt is as follows:
\begin{promptbox}
We are designing computable and interpretable features for a text quality assessment task. The task scenario is as follows:

\texttt{\{scene\_description\}}

Now, please fully embody the following role and think from its perspective. The role and its evaluation angle are: \texttt{\{role\_description\}}.
Your task is to propose a set of candidate features for measuring text quality based on your role and evaluation criteria. These features should be concrete, measurable, and interpretable. The feature description must be clear enough for someone to understand how to evaluate the text based on it.

Output \texttt{\{feature\_count\_per\_role\}} of what you consider the most important features. Each feature must be on a new line, and should be about 30 words.
Only output \texttt{\{feature\_count\_per\_role\}} lines, with each line containing a distinct feature. For each feature, provide only its name and a clear text description of what it measures. Do not use any special symbols or formatting, including list numbers.

Example Format:

{[Feature 1]}, {[Feature description]}

{[Feature 2]}, {[Feature description]}
\end{promptbox}

\paragraph{Contrastive Analysis Prompts} Used to generate features by comparing high and low-quality sample texts.

\subparagraph{Analyze Positive Samples Prompt.} Used to analyze common features in high-quality texts.
\begin{promptbox}
We are designing features for a text quality assessment task. The task scenario is as follows:

\texttt{\{scene\_description\}}

We have selected a batch of texts identified as high-score under a certain evaluation system. Here are some of those samples:

\texttt{\{samples\}}

Your task is to carefully analyze these high-score text samples and summarize the common features they possess that could explain their high scores. These features should be concrete, measurable, and interpretable. The feature description must be clear enough for someone to understand how to evaluate the text based on it.

Output \texttt{\{feature\_count\_positive\}} most important features. Each feature must be on a new line, and should be about 30 words.
Only output \texttt{\{feature\_count\_positive\}} lines, with each line containing a distinct feature. For each feature, provide only its name and a clear text description of what it measures. Do not use any special symbols or formatting, including list numbers.

Example Format:

{[Feature 1]}, {[Feature description]}

{[Feature 2]}, {[Feature description]}
\end{promptbox}

\subparagraph{Analyze Negative Samples Prompt.} Used to analyze common features in low-quality texts.
\begin{promptbox}
We are designing features for a text quality assessment task. The task scenario is as follows:

\texttt{\{scene\_description\}}

We have selected a batch of texts identified as low-score under a certain evaluation system. Here are some of those samples:

\texttt{\{samples\}}

Your task is to carefully analyze these low-score text samples and summarize the common features they possess that could explain their low scores. These features should be concrete, measurable, and interpretable. The feature description must be clear enough for someone to understand how to evaluate the text based on it.

Output \texttt{\{feature\_count\_negative\}} most important features. Each feature must be on a new line, and should be about 30 words.
Only output \texttt{\{feature\_count\_negative\}} lines, with each line containing a distinct feature. For each feature, provide only its name and a clear text description of what it measures. Do not use any special symbols or formatting, including list numbers.

Example Format:

{[Feature 1]}, {[Feature description]}

{[Feature 2]}, {[Feature description]}
\end{promptbox}

\subparagraph{Analyze Contrastive Samples Prompt.} Used for contrastive analysis between high and low-quality texts.
\begin{promptbox}
We are designing features for a text quality assessment task. The task scenario is as follows:

\texttt{\{scene\_description\}}

We have selected batches of texts identified as high-score and low-score under a certain evaluation system.

High-Score Samples:

\texttt{\{positive\_samples\}}

Low-Score Samples:

\texttt{\{negative\_samples\}}

Your task is to conduct a contrastive analysis of the two sample sets and summarize the most significant distinguishing features. These should be features that high-score texts possess and low-score texts lack. The features must be concrete, measurable, and interpretable. The feature description must be clear enough for someone to understand how to evaluate the text based on it.

Output \texttt{\{feature\_count\_contrastive\}} of the most distinctive features. Each feature must be on a new line, and should be about 30 words.
Only output \texttt{\{feature\_count\_contrastive\}} lines, with each line containing a distinct feature. For each feature, provide only its name and a clear text description of what it measures. Do not use any special symbols or formatting, including list numbers.

Example Format:

{[Feature 1]}, {[Feature description]}

{[Feature 2]}, {[Feature description]}
\end{promptbox}

\paragraph{Integrate Features Prompt.} Used to consolidate and deduplicate the generated feature candidates. This prompt is as follows:
\begin{promptbox}
We have generated a batch of candidate features for text quality assessment through various methods (multi-role perspectives, data sample analysis, etc.). They now need to be consolidated.

Original Feature List:

\texttt{\{feature\_list\}}

As a feature engineering expert, your task is to process the original feature list above to produce a final, refined pool of candidate features.

Processing Requirements:

1. Merge and Deduplicate: Identify and merge features that are semantically identical or highly similar.

2. Optimize Descriptions: Ensure each feature's description is clear, precise, unambiguous, and actionable for the subsequent development of annotation tools.

3. Format Output: Organize the output into a clean list.

Output each feature on a new line. For each feature, provide only a detailed text description of what it measures.
The final list should contain as many unique features as can be derived from the original list after processing.
Just output a plain list of features. Do not use any special symbols or formatting, including list numbers.
Start a new line ONLY when moving to the next feature. If you find n features, just output n lines, with each line containing a distinct feature.
\end{promptbox}

\subsubsection{Autonomous Tool Implementation}
These prompts are used for the autonomous tool implementation phase (Section~\ref{ssec:tool_implementation}), where the agent decides on a tool type and generates the corresponding code or annotation prompt.

\paragraph{Decide Tool Type Prompt.} Used to determine whether a feature is best measured by code or an LLM prompt. This prompt is as follows:
\begin{promptbox}
Your task is to determine the best tool type to annotate a text feature. The options are ``CODE'' or ``PROMPT''.

``CODE'' is for features that are simple, explicit, and can be accurately auto-annotated with Python code without requiring intelligent processing.

``PROMPT'' is for features that cannot be accurately annotated with simple rule-based code, and may be abstract, nuanced, subjective, or require deep semantic understanding.

Feature Description:

\texttt{\{feature\_description\}}

Based on the description, is this feature better suited for ``CODE'' or ``PROMPT''?
Respond with a single word: either ``CODE'' or ``PROMPT''. Do not provide any other text or explanation.
\end{promptbox}

\paragraph{Generate Code Tool Prompt.} Used to generate a self-contained Python function for annotation.
\begin{promptbox}
Your task is to write a Python function that serves as an annotation tool for a specific text feature.

The function should:

1. Be named \texttt{\{function\_name\}}.

2. Accept a single string argument named \texttt{text}.

3. Return a single numerical value (float or int).

4. Be self-contained. You can use common libraries like \texttt{re}, \texttt{nltk}, \texttt{textblob}, but do not assume any external files are available.

5. If a library is used, include the necessary import statement inside the function to ensure it's encapsulated.

Here is the feature the function needs to measure:

Feature: \texttt{\{feature\_name\}}

Description: \texttt{\{feature\_description\}}

Generate the complete Python code for this function. Do not include any text or explanation outside the function's code block. Start the response directly with the function definition.

Example:
\begin{verbatim}
def annotate(text: str) -> float:
    # import necessary libraries here
    # ... function logic ...
    return score
\end{verbatim}
\end{promptbox}

\paragraph{Generate Prompt Tool Prompt.} Used to generate an LLM prompt for annotating complex features. This prompt is as follows:
\begin{promptbox}
Your task is to create an precise and effective prompt template for a Large Language Model to use as a feature annotation tool.

This template will be used to evaluate different pieces of text. It must contain the placeholder \texttt{[TEXT\_TO\_EVALUATE]} where the actual text will be inserted later.

The prompt you create should instruct the LLM to:

1. The LLM should evaluate the text based on the feature described below.

2. Provide a numerical score on a scale of 1 to 10 (where 1 is low quality/absence of the feature, and 10 is high quality/strong presence of the feature).

3. Respond with ONLY the numerical score, without any additional text or explanation.

4. Clearly explain the criteria used to determine the score.

Here is the feature that the annotation prompt needs to measure:

\texttt{\{feature\_description\}}

Now, generate the annotation prompt template text. Your ONLY task is to generate the raw text for a prompt template. Do not output anything else. Do not use markdown, do not add titles, do not add any explanations.

Your output must begin directly with the text of the prompt. Your output should end with ``The text to evaluate is: \texttt{[TEXT\_TO\_EVALUATE]}.'' The \texttt{[TEXT\_TO\_EVALUATE]} placeholder should only be used once at the end of the prompt.
\end{promptbox}

\subsubsection{Reflective Search and Memory}
These prompts support the reflective search and dual-level memory architecture (Sections~\ref{ssec:reflective_search} and~\ref{ssec:memory}).

\paragraph{Reflect and Generate Features Prompt (Intra-Task Reflection).} Guides the agent to reflect on its progress and hypothesize new features mid-task. This prompt is as follows:
\begin{promptbox}
You are an expert in text quality assessment feature engineering. We are working on a text quality assessment task with the following scenario:

Scenario Description:

\texttt{\{scene\_description\}}

We have completed a round of feature selection using a beam search algorithm. The current best features selected and their performance (Spearman correlation coefficients, sorted in descending order) are:

\texttt{\{features\_with\_scores\}}

Your task is to analyze the selected features and their performance in the context of the above scenario, then propose \texttt{\{new\_feature\_count\}} NEW, innovative features that could potentially improve the assessment quality for this specific task.

When designing new features, you should:

1. Consider the specific requirements and context of the scenario described above

2. Identify potential gaps or aspects not well-covered by the current feature set

3. Consider features that might capture different dimensions or perspectives relevant to this scenario

4. Think about features that could complement or enhance the existing ones

5. Ensure the features are concrete, measurable, and interpretable

6. Avoid proposing features that are too similar to existing ones

Output exactly \texttt{\{new\_feature\_count\}} new features. Each feature must be on a new line, and should be about 30 words.

Only output \texttt{\{new\_feature\_count\}} lines, with each line containing a distinct feature. For each feature, provide only its name and a clear text description of what it measures. Do not use any special symbols or formatting, including list numbers.

Example Format:

[Feature 1], [Feature description]

[Feature 2], [Feature description]
\end{promptbox}

\paragraph{Cross-Scene Learning Prompt (Cross-Task Memory).} Enables the agent to use memories from past tasks to bootstrap hypothesis generation for a new task. This prompt is as follows:
\begin{promptbox}
You are an expert in text quality assessment feature engineering. We are working on a NEW text quality assessment task. Your goal is to learn from the successful features identified in OTHER similar tasks and propose new features for the current task.

Current Task Scenario:

\texttt{\{current\_scene\_description\}}

We have analyzed similar text quality assessment tasks from different domains. Here are the successful features discovered in those tasks:

\texttt{\{other\_scenes\_info\}}

Based on the successful patterns and features from these other tasks, please propose \texttt{\{feature\_count\}} NEW, innovative features that could be effective for the CURRENT task described above.

When designing new features, you should:

1. Identify common patterns across the successful features from other tasks

2. Adapt and generalize these patterns to fit the current task's context

3. Consider what made those features successful and how similar principles apply here

4. Ensure the features are concrete, measurable, and interpretable

5. Make sure the features are relevant to the current task scenario

Output exactly \texttt{\{feature\_count\}} new features. Each feature must be on a new line, and should be about 30 words.
Only output \texttt{\{feature\_count\}} lines, with each line containing a distinct feature. For each feature, provide only its name and a clear text description of what it measures. Do not use any special symbols or formatting, including list numbers.

Example Format:

[Feature 1], [Feature description]

[Feature 2], [Feature description]
\end{promptbox}

\end{document}